\documentclass[letterpaper, 10 pt, conference]{ieeeconf}  

\IEEEoverridecommandlockouts                              

\usepackage{amsmath} 
\usepackage{amssymb}  
\usepackage{algorithm}
\usepackage[noend]{algpseudocode}
\usepackage{color}
\usepackage{graphicx}
\usepackage{booktabs}

\title{\LARGE \bf Learning Locomotion Controllers for Walking Using Deep FBSDE}

\author{Bolun Dai$^{1}$, Virinchi Roy Surabhi$^{1}$, Prashanth Krishnamurthy$^{1}$, Farshad Khorrami$^{1}$
\thanks{$^{1}$Control/Robotics Research Laboratory, Electrical~\&~Computer Engineering Department, Tandon School of Engineering (Polytechnic Institute), New York University, Brooklyn, NY, 11201
{\tt\small bd1555@nyu.edu, virinchi.roy@nyu.edu, prashanth.krishnamurthy@nyu.edu, khorrami@nyu.edu}
}
}

\begin{document}
\maketitle
\thispagestyle{empty}
\pagestyle{empty}

\maketitle
\begin{abstract}
In this paper, we propose a deep forward-backward stochastic differential equation (FBSDE) based control algorithm for locomotion tasks. We also include state constraints in the FBSDE formulation to impose stable walking solutions or other constraints that one may want to consider (e.g., energy). Our approach utilizes a deep neural network (i.e., LSTM) to solve, in general, high-dimensional Hamilton–Jacobi–Bellman (HJB) equation resulting from the stated optimal control problem. As compared to traditional methods, our proposed method provides a higher computational efficiency in real-time; thus yielding higher frequency implementation of the closed-loop controllers. The efficacy of our approach is shown on a linear inverted pendulum model (LIPM) for walking. Even though we are deploying a simplified model of walking, the methodology is applicable to generalized and complex models for walking and other control/optimization tasks in robotic systems. Simulation studies have been provided to show the effectiveness of the proposed methodology.
\end{abstract}
\section{Introduction}
How to enable robots to perform locomotion tasks is an important problem in robotics research given the wide applications of walking robots. Walking robots are able to move without the existence of roads or other types of infrastructures. Their legs makes it possible to move with small support surfaces~\cite{7798370} making walking robots more appealing for tasks performed on rugged terrains~\cite{5509176}. Also,  movement of legged robots does not rely on the existence of air or propellants, which makes it an ideal candidate for extraterrestrial expeditions~\cite{doi:10.5772/58731}. Additionally, the human-like figure and motion~\cite{4209468}~\cite{1248841} makes it easier to integrate humanoids into our society. However, given the complexity, the control of walking robots has been a challenging task. 

Traditionally, model predictive control (MPC) has been widely used in locomotion tasks~\cite{7353843}~\cite{7384453}~\cite{8624921}. The authors in~\cite{5654429} proposed a complete solution for walking on flat ground, which includes turning. The authors in~\cite{7363423} then expanded on previous works to enable walking in 3D, e.g. walking up stairs. At each time step, MPC-based methods rely on optimization solvers~\cite{cvx} to solve for control at a given time considering the dynamics over a preview horizon. This process can be computationally intensive especially for high dimensional systems over long preview horizons.

To mitigate computational issues, deep learning based approaches are being applied in locomotion tasks. Compared to the resources required for solving a convex optimization problem, the computational cost is much less for one forward pass through a neural network. Recently, deep reinforcement learning has been widely applied to locomotion tasks. The role  deep reinforcement learning plays is versatile, it can be used in conjunction with a feedback controller where the learned policy provides a reference state~\cite{8793864}, or provide set points to a motor PD controller~\cite{Siekmann2020LearningMC}. However, deep reinforcement learning methods require a well-designed-task-specific reward function, which in the case of locomotion is hard to obtain. The reward functions in both~\cite{8793864} and~\cite{Siekmann2020LearningMC} look rather arbitrary and require exhaustive tuning to find the best balance between different design objectives. A different line of deep learning based approaches attempts to mimic the behavior of existing optimization-based controllers in a supervised learning fashion~\cite{DBLP:journals/corr/abs-2011-02818}. However, this approach requires access to existing optimization-based controllers or a dataset of state-control pairs, which is not always available.

Locomotion problems are essentially optimal control problems. One method that utilizes deep learning to solve optimal control problems is deep FBSDE. The idea behind deep FBSDEs is to find a numerical approximation to the solution of the HJB equation, which is the value function given with respect to a cost function. Following the derivation in~\cite{DBLP:conf/rss/WangPT19}, the optimal control can then be calculated as a function of the partial derivative of the value function with respect to the state. We can formulate this problem as a FBSDE and solve it using a neural network which has the benefit of resolving compounding-errors. Such a method has gained traction recently. \cite{DBLP:conf/rss/WangPT19} first proposed the deep FBSDE algorithm using LSTMs. On top of the vanilla LSTM-based FBSDE formulation,~\cite{DBLP:journals/corr/abs-1906-04762} showed how to handle systems with control multiplicative noise and \cite{DBLP:conf/cdc/WangLPET19} solved problems with unknown noise distribution. In our recent work \cite{acc2021}, we expanded deep FBSDEs to consider problems with state constraints. However, these works only considered examples with continuous dynamics.

The present paper aims to expand the usage of deep FBSDE controllers to locomotion tasks, which is a hybrid dynamical system. In Section II, the walking model LIPM is introduced and the problem formulation is given. In Section III, an outline of the deep FBSDE controller is provided. Section IV shows the methodology for deep FBSDE controller to adapt to walking tasks, and changes to the neural network architecture are highlighted. In Section V, we demonstrate the efficacy of our approach on LIPM model, along with its state constrained version, in simulation. We also compare the computation efficiency of our method with CVX. Finally, in Section VI, the paper is concluded and a brief discussion on extensions of this work is presented.
\section{Problem Formulation}
In this section, we discuss key aspects of the LIPM walking problem. The LIPM, under the assumption of no jumping and no large rotation movements in the sagittal plane, is given as
\begin{equation}
    \mathbf{c}^{x, y} - \frac{c^z}{g^z}\Ddot{\mathbf{c}}^{x, y} = \mathbf{p}^{x, y},
    \label{eq:lipm_model}
\end{equation}
where $\mathbf{c}^{x, y} = [c^x\ c^y]^T$ denotes the position of the center of mass (CoM) in the $xy$-plane (transverse plane), $c^z$ is the height of the CoM, $g^z$ is the vertical component of the gravitational acceleration, $\Ddot{\mathbf{c}}^{x, y} = [\Ddot{\mathbf{c}}^x\ \Ddot{\mathbf{c}}^x]^T$ denotes the acceleration of the CoM in the $xy$-plane and $\mathbf{p}^{x, y} = [p^x\ p^y]^T$ denotes the position of the center of pressure (CoP) in the $xy$-plane. If we treat the CoP location $\mathbf{p}^{x, y}$ as the control input we can write the dynamical system as
\begin{equation}
    \Ddot{\mathbf{c}}^{x, y} = \frac{g^z}{c^z}\mathbf{c}^{x, y} - \frac{g^z}{c^z}\mathbf{p}^{x, y},
    \label{eq:lipm_cont}
\end{equation}
which has the form of $\dot{x} = Ax + Bu$. In matrix form, equation~\eqref{eq:lipm_cont} can be written as
\begin{equation}
    \dot{\bar{\mathbf{c}}} = \begin{bmatrix}
        \mathbf{0}_{2\times2} & I_{2\times2}\\
        g^z/c^zI_{2\times2} & \mathbf{0}_{2\times2}
    \end{bmatrix}\bar{\mathbf{c}} + \begin{bmatrix}
        \mathbf{0}_{2\times2}\\
         g^z/c^zI_{2\times2}
    \end{bmatrix}\mathbf{p}^{x, y},
    \label{eq:state_space_lipm}
\end{equation}
with $\bar{\mathbf{c}} = [c^x\ c^y\ \Dot{c}^x\ \Dot{c}^y]^T$.

The purpose of this work is to achieve stable walking behavior, thus the phases when the robot starts from standing still and coming to a stop are not considered. Starting from a CoM state $\bar{\mathbf{c}}_0 = \xi$ that has an initial velocity, we want to build a controller that is able to maintain stable walking over an arbitrary long time period. We use the cost function
\begin{subequations}
\begin{align}
    \min_{\mathbf{p}_i^{x, y}} &\sum_{i=0}^{N-1}{\Big[q(\bar{\mathbf{c}_i}) + r(\mathbf{p}_i^{x, y})\Big]} + g(\bar{\mathbf{c}}_N)\\
    &\bar{\mathbf{c}}_{i+1} = \bar{\mathbf{c}}_i + \dot{\bar{\mathbf{c}}}\Delta{t}\\
    &\bar{\mathbf{c}}_0 = \xi\\
    &-\frac{1}{2}l_x \leq p^x \leq \frac{1}{2}l_x\\
    &-\frac{1}{2}l_y \leq p^y \leq \frac{1}{2}l_y,
\end{align}
\label{eq:control_problem}
\end{subequations}
\noindent where $q(\bar{\mathbf{c}_i})$ is the state cost, $r(\mathbf{p}_i^{x, y})$ is the control cost, $g(\bar{\mathbf{c}}_N)$ is the terminal state cost and $l_x$ and $l_y$ denotes the length of the foot along the $x$, $y$ direction. The inequality constraints denotes the CoP has to be within the support polygon.
\section{Deep FBSDE}
\begin{figure*}[t!]
    \centering
    \includegraphics[width=\textwidth]{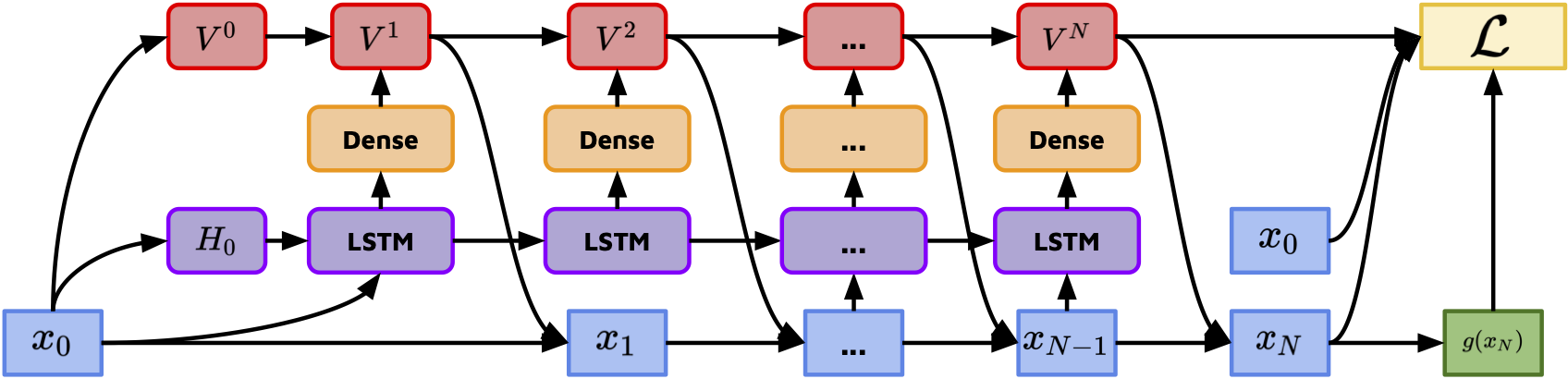}
    \caption{\textbf{Neural Network Architecture}: the architecture here is for $N$ timesteps, the connection between $V^n$ and $x_n$ denotes applying control actions, the weights are shared among each time step. Curve edged boxes represents trainable parameters, sharp edged boxes represents intermediate calculated values.}
    \label{fig:nn_arch}
\end{figure*}
In this section, we outline the design of the deep FBSDE controller. For a system with dynamics that includes disturbances, it can be described using a stochastic differential equation (SDE),
\begin{equation}
	\label{eq:SDE}
	dx(t) = f(x(t),t)dt + G(x(t),t)u(t)dt + \Sigma(x(t),t)dw(t)
\end{equation}
with initial condition $x(0) = x_0$ and $w(t)\in\mathbb{R}^\nu$ being a standard Brownian motion. The states are denoted by $x\in\mathbb{R}^n$, the control input by $u\in\mathbb{R}^m$, and time by $0<t<T<\infty$. In \eqref{eq:SDE}, $f: \mathbb{R}^n\times[0, T]\rightarrow\mathbb{R}^n$ represents the drift, $G: \mathbb{R}^n\times[0, T]\rightarrow\mathbb{R}^{n\times m}$ represents the control influence, and $\Sigma: \mathbb{R}^n\times[0, T]\rightarrow\mathbb{R}^{n\times\nu}$ represents the diffusion (influence of the Brownian motion to the state). The purpose of the deep FBSDE controller is to find $u(t)$ that minimizes an objective function $J^u(x,t)\in\mathbb{R}^+$. The objective function is denoted as
\begin{equation}
    J^u(x,t) = \mathbb{E}\Big[g(x(T)) + \int_t^T\Big(q(x(s)) +\frac{1}{2}u(s)^TRu(s)\Big)ds\Big]
    \label{eq:objective_optimization}
\end{equation}
where $g:\mathbb{R}^n\rightarrow\mathbb{R}^+$ is the terminal state cost, $q:\mathbb{R}^n\rightarrow\mathbb{R}^+$ is the instantaneous state cost, and the control cost matrix is $R\in\mathbb{R}^{m\times m}$, which is positive definite. The value function $V(x,t)\in\mathbb{R}^+$ is defined as
\begin{equation}
    V(x,t) := \inf J^{u}(x,t)
\end{equation} 
where the $\inf$ is computed over all control signals $u(.)$ over the time interval $(t, T]$.

Following derivations in~\cite{acc2021}, the HJB equation is given as
\begin{subequations}
\begin{align}
    & V_t(x,t) + \mathcal{L}V(x,t) + h(x,V_x,t)\\
    & V(x,T) = g(x).
\end{align}
\label{eq:HJB}
\end{subequations}
The generator function $\mathcal{L}V(x,t)$ is defined as
\begin{equation}
    \mathcal{L}V = \frac{1}{2}\hbox{trace}(\Sigma\Sigma^TV_{xx}) + f^TV_x
\end{equation}
with $V_x$ and $V_{xx}$ representing the first and second order partial derivative of $V$ w.r.t. $x$, respectively. The Hamiltonian is defined as
\begin{equation}
    h(x,V_x,t) \!=\! \inf_{u\in\mathcal U}\left(q(x)+(G(x,t)u)^TV_x(x,t)+\frac{1}{2}u^TRu\right),
\end{equation}
with $\mathcal{U}$ being the admissible set of controls. By taking the minimum of the Hamiltonian we can analytically compute the optimal control as a function of $V_x$, the partial derivative of the value function with respect to the state,
\begin{equation}
	\label{eq:feedbackForm}
	u^*(x,t) = -R^{-1}G^T(x,t)V_x(x,t).
\end{equation}
We can solve for~\eqref{eq:feedbackForm} using a deep FBSDE
\begingroup
\allowdisplaybreaks
\begin{subequations}
\label{eq:FBSDE_dnn}
\begin{align}
dy(t) =&\ \Big(-h\big(x(t), \Sigma^T(x(t),t)V_x(x(t),t; \theta), t\big)\nonumber\\
        &+ V_x^T\big(x(t), t; \theta\big)\Sigma\big(x(t),t\big)\Gamma\big(x(t),t\big)u\big(x(t),t\big)\Big)dt\nonumber\\ 
        &+ V_x^T\big(x(t),t; \theta\big)\Sigma\big(x(t),t\big)dw(t)\\
dx(t) =&\ \Big(f\big(x(t),t\big) +\Sigma\big(x(t),t\big)\Gamma\big(x(t),t)u(x(t),t\big)\Big)dt\nonumber\\
       &+ \Sigma\big(x(t),t\big)dw(t)\\
u(t) =&\ -R^{-1}\Gamma^T\big(x(t), t\big)\Sigma^T\big(x(t), t\big)V_x(x(t),t; \theta)\\
y(0) =&\ V\big(\phi, x_0\big)\\
x(0) =&\ x_0,
\end{align}
\end{subequations}
where $y(t) = V(x_t) = V^t$ and $G(x(t), t) = \Sigma(x(t), t)\Gamma(x(t), t)$,  $\Gamma: \mathbb{R}^n\times[0, T]\rightarrow\mathbb{R}^{\nu\times m}$. This shows the range of $G$ is a subset of the range of $\Sigma$, which excludes the case of a channel containing control input without noise. The neural network that estimates $V_x$ at each time step is parameterized by $\theta$ and the neural network that estimates the value function at the initial time step $V^0$ is parameterized by $\phi$. The details of the neural network architecture will be explained in the following sections.

\section{Controller Design}
Unlike canonical control examples shown in~\cite{acc2021}, the nature of locomotion tasks creates a hybrid dynamical system, where the dynamics is discontinuous between two steps. Additionally, stable locomotion behavior usually creates a cyclic movement, where a region of attraction exists for the terminal state at each step (will discuss in detail in the next section). Due to these unique conditions, instead of training the controller over multiple steps, the learning problem will be ``normalized" if we only focus on one single step.

Using this intuition, the learned controller will be only trained using experience from one foot step. Experiences obtained from one foot step can have low variance, thus noise is added to the initial state of each step in addition to the disturbances injected to state dynamics. Therefore, instead of having a single initial state as in~\cite{acc2021}, the initial state will be sampled from a distribution. To accommodate this change, the initial value function is now a function of the initial state. For LSTMs, the initial hidden states $H_0$, parameterized by trainable weights, also takes the initial state as input. The overall neural network architecture is shown in Figure~\ref{fig:nn_arch}. The loss function is defined as
\begin{equation}
    L = \frac{1}{M}\sum_{m=1}^{M}\Big[\|g(x_N^m) - y_N^m\|_2^2 + \lambda_1\|x_0^m - x_N^m\|_2^2\Big] + \lambda_2\|\theta\|_2^2,
\end{equation}
where $M$ denotes the batch size, and $\lambda_1$ and $\lambda_2$ determines the weight of the stability enforcement term and the regularization term, respectively. Compared to~\cite{acc2021}, the additional term in the loss function is designed to encourage a cyclic motion.
\section{Experiment}
\begin{figure*}[t!]
    \centering
    \includegraphics[width=\textwidth]{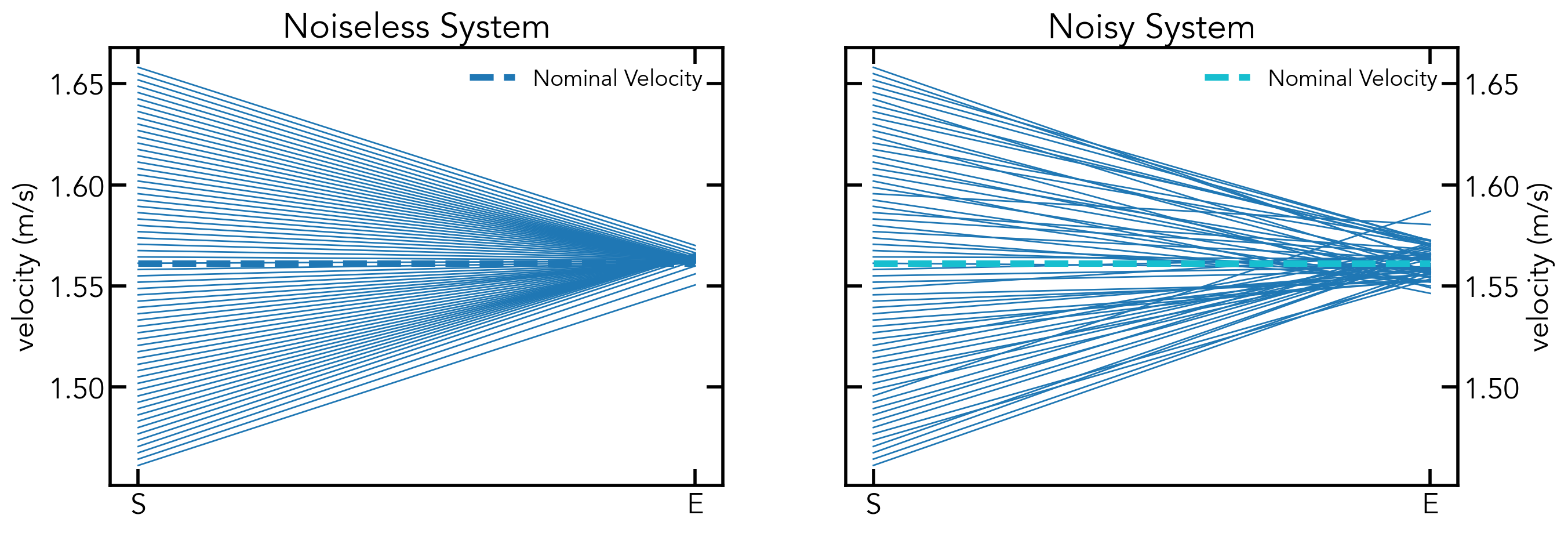}
    \caption{\textbf{Training Results}: these two figures demonstrates the stability of the learned policy. The ``S" on the $x$-axis represents the start of each foot step and ``E" represents the end of each foot step. In both of the plots there are 64 pairs. The value of each line on the ``S" side is the velocity of the CoM at the start of each foot step, the value on the ``E" side is the velocity of the CoM at the end of each foot step. The thick dashed line in the middle is a nominal velocity that serves as a target during training. Since $x$ and $y$ motion are decoupled only results of the $x$ direction are shown.}
    \label{fig:trainResults}
\end{figure*}

\begin{figure*}[t!]
    \centering
    \includegraphics[width=\textwidth]{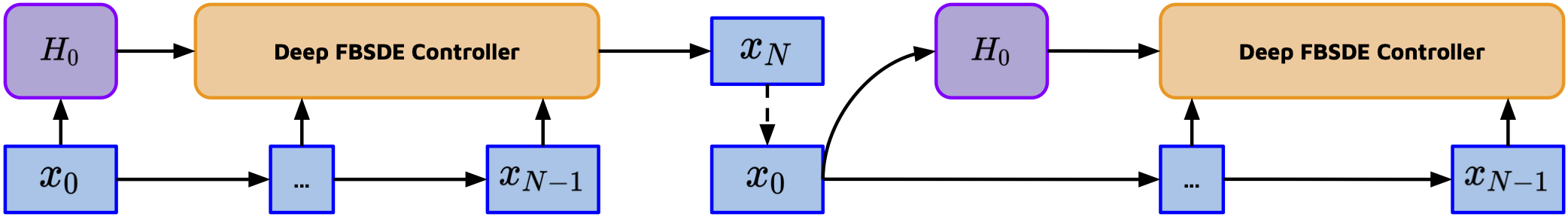}
    \caption{\textbf{Workflow of walking multiple steps}: this figure illustrates how to used the learned deep FBSDE controller trained on a single footstep to walk multiple steps. The round edge boxes denotes trainable components, while the sharp edge boxes denotes intermediate values. The dashed arrow in the middle denotes the change in coordinate systems.}
    \label{fig:stableWalking}
\end{figure*}

In this section, we show the efficacy of our approach on the LIPM model, with both noiseless and noise injected state dynamics. First, we introduce the setup of the experiments, then the training results are shown. Second, we show this learned policy trained on a single step is able to generalize to stable walking over multiple steps. Third, we demonstrate the capability of our method on handling state constraints. Finally, we compare the computation efficiency between our approach and traditional optimization based methods. All experiments are implemented using Tensorflow.

\subsection{Experiment Setup}
To align the formulation of~\eqref{eq:state_space_lipm} and~\eqref{eq:SDE} we have
\begin{subequations}
\begin{align}
    f &= \begin{bmatrix}
        \mathbf{0}_{2\times2} & I_{2\times2}\\
        g^z/c^zI_{2\times2} & \mathbf{0}_{2\times2}
    \end{bmatrix}\\
    g &= \begin{bmatrix}
        \mathbf{0}_{2\times2}\\
         g^z/c^zI_{2\times2}
    \end{bmatrix}\\
    x &= \bar{\mathbf{c}}.
\end{align}
\end{subequations}
Disturbances are only injected to the acceleration terms, thus $\Sigma = \mathrm{diag}[0\ 0\ 1\ 1]$. The CoM height $c^z$ is fixed at $1m$ and $g^z = -9.81m/s^2$. We use quadratic costs for the state cost, controls cost, and terminal cost defined in~\eqref{eq:control_problem}
\begin{subequations}
\begin{align}
    q(\bar{\mathbf{c}_i}) &= \mathbf{c}_i^TQ_i\mathbf{c}_i\\
    r(\mathbf{p}_i^{x, y}) &= (\mathbf{p}_i^{x, y})^TR_i\mathbf{p}_i^{x, y}\\
    g(\bar{\mathbf{c}}_N) &= \bar{\mathbf{c}}_N^TQ_N\bar{\mathbf{c}}_N,
\end{align}
\end{subequations}
where
\begin{subequations}
\begin{align}
    Q_i &= \mathrm{diag}[1\ 1\ 1\ 1]\\
    R_i &= \mathrm{diag}[100\ 100]\\
    Q_N &= \mathrm{diag}[100\ 100\ 100\ 100].
\end{align}
\end{subequations}
The robot has a foot with the size of $0.15 m\times 0.15 m$. The foot step position are pre-planned to walk along the $x=y$ line. The duration of each footstep is $0.7s$, the step size is $0.8\sqrt{2}m$, which corresponds to $0.8m$ in both the positive $x$ and $y$ axis. When simulating the state dynamics we use a time step of $0.02s$. In training, we use a learning rate of 1e-3 over 10,000 iterations. To enhance the robustness of the learned policy, a uniformly sampled disturbance is applied to the initial state on both systems with noiseless and noisy dynamics.

\subsection{Training Results}
To analyze the quality of learned policies, we treat the relationship between the velocity of the initial state of a foot step $v_\mathrm{S}$ and the velocity of the terminal state $v_\mathrm{E}$ as a mapping. Stable walking behavior can be observed when this mapping maps an arbitrary $v_\mathrm{S}$ that is in the vicinity of the nominal velocity to a fixed $v_\mathrm{E}$ or to a small $v_\mathrm{E}$ interval. Physically, this is equivalent to showing under disturbances, the system is able to adjust the subsequent control actions to drag the CoM velocity back to a desired range. The mapping for both the noiseless and noisy systems are shown in Figure~\ref{fig:trainResults}. For the noiseless system, a clear region of attraction can be seen: the $v_S$'s are map to a very small interval surrounding the nominal velocity. Though not equal to the nominal velocity, this clearly shows robustness of the learned policy. For the noisy system, such a convergence pattern is also present, though larger than the noiseless case, the $v_S$'s are also mapped to a small interval around the nominal velocity.

\subsection{Walking}
\begin{figure*}[t!]
    \centering
    \includegraphics[width=\textwidth]{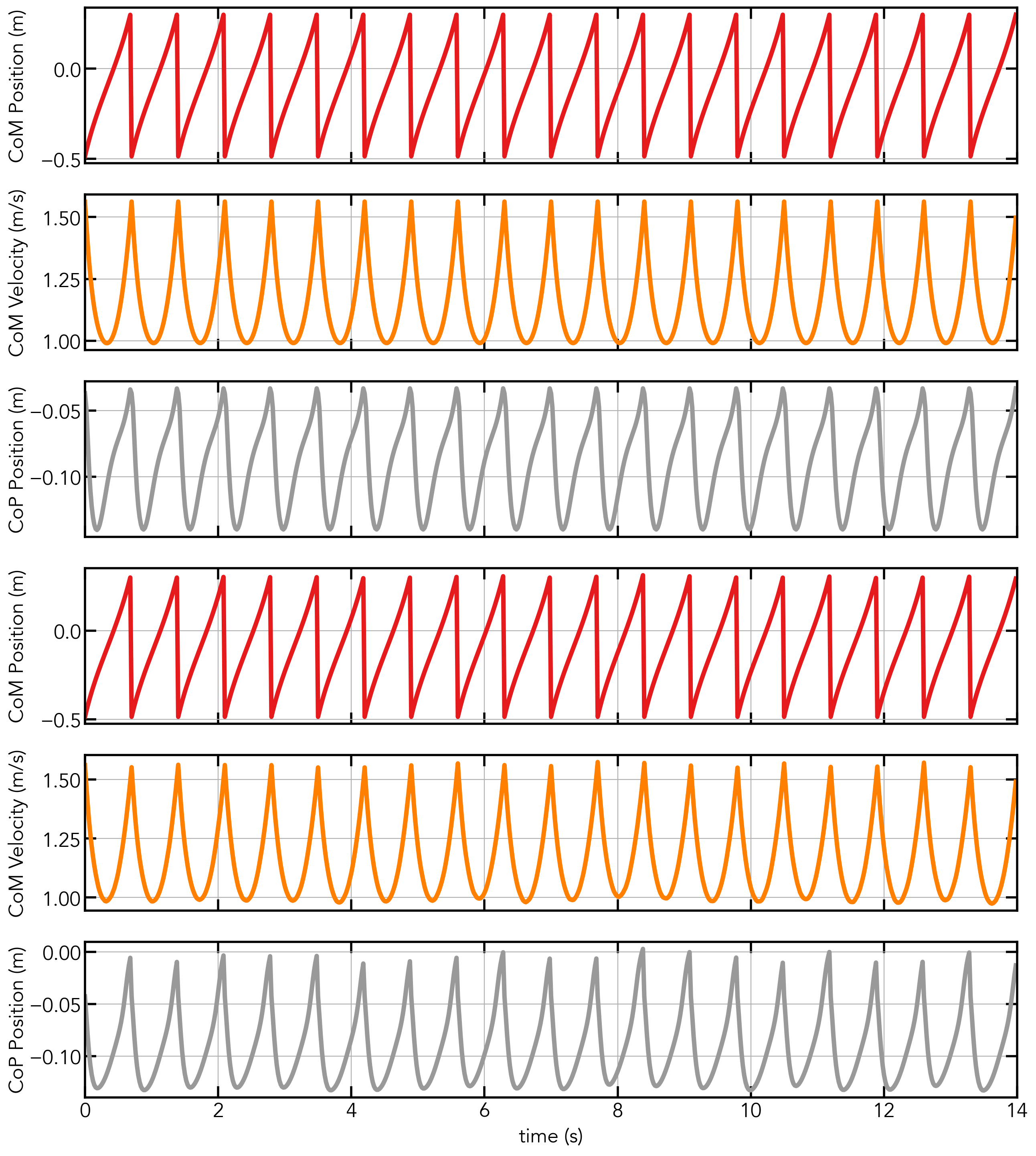}
    \caption{\textbf{Walking for 20 steps using the learned policy}: the top three figures shows the evolution of the CoM and CoP over 20 steps using a noiseless system, the bottom three shows the evolution of its noisy counterpart. The CoP is clipped between $[-0.15, 0.15]$. All measurements are performed inside a local coordinate system that is attached to the center of the foot.}
    \label{fig:stableWalkingDemo}
\end{figure*}

In this section, we demonstrate how policy trained on a single foot step can be utilized for walking over multiple steps. The learned policy uses states defined in a local frame, where the origin is at the center of the foot, and its $x$ and $y$ axis are aligned with the global frame. At the transition between two foot steps, the states are updated to be measured in the local body frame of the next foot step. When updating the states, the CoM velocities stays fixed (velocities are the same when observing from static frames). However, we can choose how big a step length to take, which changes the CoM position in the body frame. We have tested using a fixed step size. However, for the learned policy, especially under the presence of disturbances, this may cause issues. If the velocity of the terminal state is slower than desired, a fixed step size might correspond to talking too large of a step. Instead, we made the initial position of the CoM of each step a fixed value $(\bar{x}_0, \bar{y}_0)$, which corresponds to a dynamic step length. The update rule is
\begin{equation}
    \begin{bmatrix}
        x_N & y_N & \dot{x}_N & \dot{y}_N
    \end{bmatrix}^T\rightarrow\begin{bmatrix}
        \bar{x}_0, \bar{y}_0 & \dot{x}_N & \dot{y}_N
    \end{bmatrix}^T.
\end{equation}
The overall workflow is shown in Figure~\ref{fig:stableWalking}. Using the proposed algorithm, the walking motion is displayed in Figure~\ref{fig:stableWalkingDemo}. As can be seen, for both the noiseless and noisy system a cyclic behavior can be maintained over 20 steps. We tested up to 40 steps, and no deterioration is shown, due to limit space only results for 20 steps are shown. 

\subsection{State Constraint}
\begin{figure}
    \centering
    \includegraphics[width=0.49\textwidth]{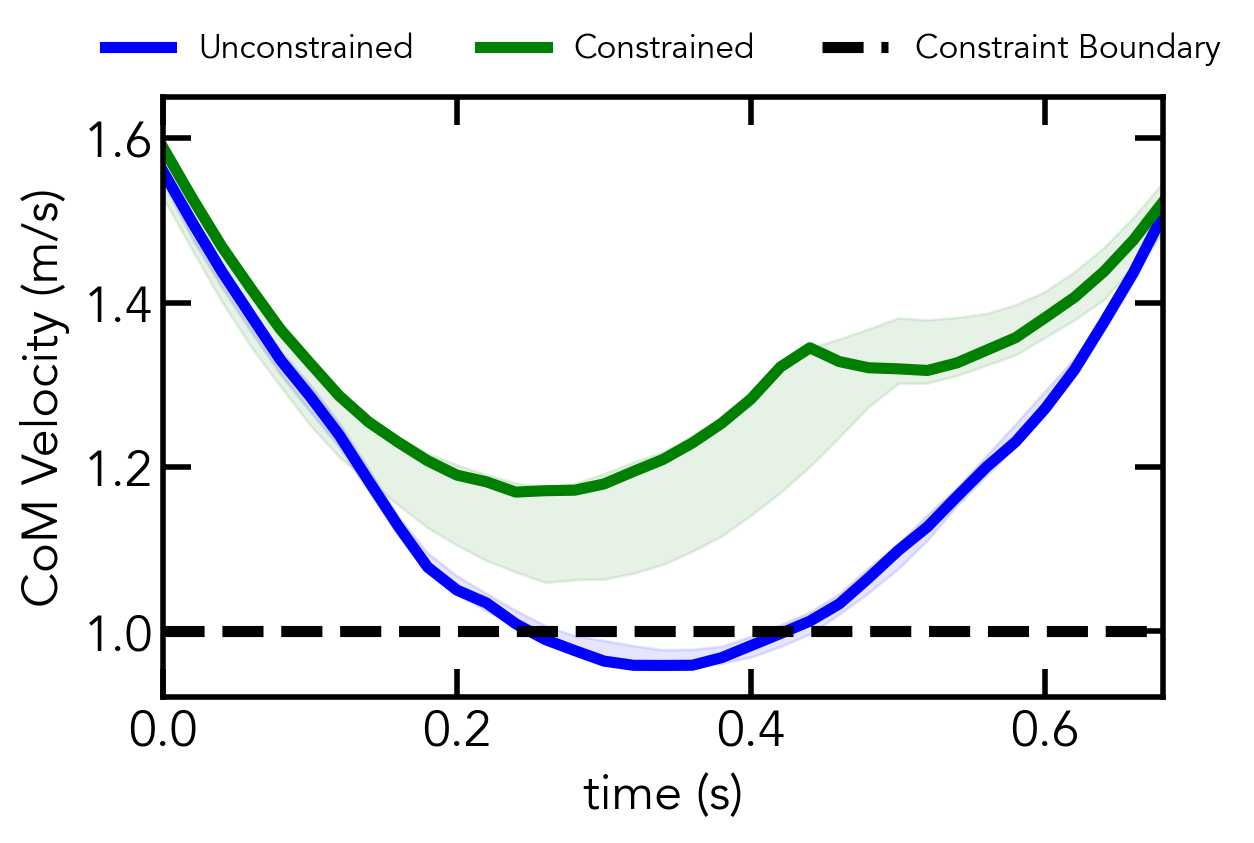}
    \caption{\textbf{Application of state constraints}: this figure compares the performance of the controller under velocity constraints. The darker blue and green trajectories are sampled trajectories among 64 consecutive steps. The light blue and green regions shows the bounds of maximum and minimum values of all the 64 trajectories at each time step. The constraint boundary is shown in black dashed lines.}
    \label{fig:constraintAnalysis2}
\end{figure}

Following the formulation in~\cite{acc2021}, a state constraint is applied to the CoM velocity for it to be faster than $1.0$m/s. We simulated our controller over 64 foot steps on a noisy model. The results for both the constrained and unconstrained setup are shown in Figure~\ref{fig:constraintAnalysis2}. We can see that for the unconstrained case every instance goes beyond the constraint boundary. For the constrained case, the constraint boundary is respected over all steps. Even though the construct of this constraint is rather artificial, it shows the capability of our method in handling state constraints.

\subsection{Computation Efficiency Comparison}
\begin{figure}
    \centering
    \includegraphics[width=0.49\textwidth]{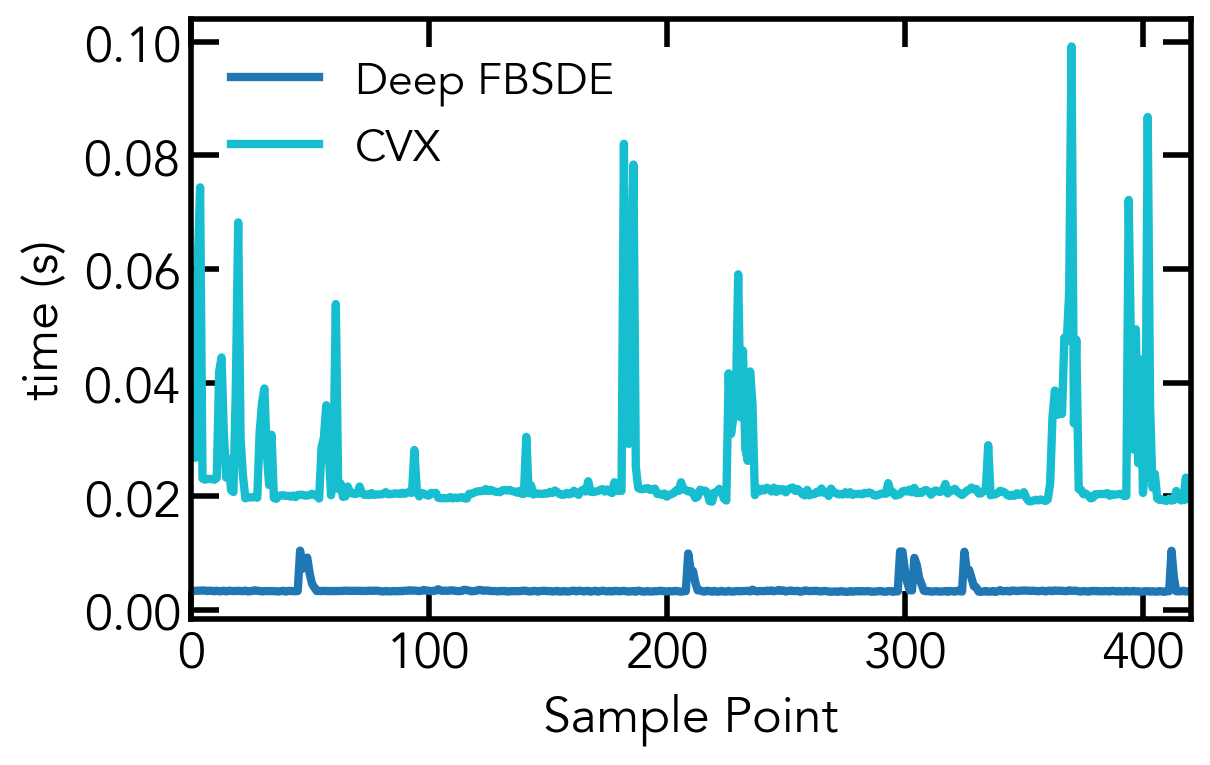}
    \caption{\textbf{Computation Efficiency Comparison}: 420 instances of control computation are sampled and the difference in computation time is displayed. In both cases, a 4 dimension state is used, the CVX solution used a preview horizon of 60 time steps. The targeted time step size is $0.02$s.}
    \label{fig:computation}
\end{figure}

In this section, we compare the computation efficiency of our approach and of using an optimization solver, in this case CVX. We compare the computation time at each time step for obtaining a control action. The comparison is shown in Figure~\ref{fig:computation}. It can be seen that our proposed method is on average 6.67 times faster and has less variance than CVX. For tasks with higher state dimensions and requires longer preview horizon this computation acceleration will be more significant. 

\section{Conclusion}
In this paper, We consider the walking problem using a simplified model. The locomotion problem of a LIPM model is solved using a deep FBSDE controller. The controller is trained on walking over a single foot step and a method is proposed to use this learned controller to generalize to walking over multiple steps. The efficacy of our approach is demonstrated in simulation on both noiseless and noisy systems. Additionally, we showed our proposed method is able to handle state constraints and is computationally much faster than optimization based methods. Potential future directions include extending to more complex models and adaptation on real hardware and its application to a full biped model.
\vspace*{-0.05in}
\bibliographystyle{IEEEtran}
\bibliography{IEEEabrv, refs.bib}
\end{document}